%% file: main.tex
\documentclass[runningheads]{llncs}

\usepackage[mobile]{eccv}

\usepackage{eccvabbrv}
\usepackage{makecell}
\usepackage{xcolor}
\usepackage{graphicx}
\usepackage{booktabs}
\usepackage{amsmath,amssymb}
\usepackage{algorithm}
\usepackage{algpseudocode}
\usepackage{multirow} 
\usepackage{pifont}
\usepackage{cite}
\usepackage{wrapfig}
\usepackage{caption} 
\usepackage[accsupp]{axessibility}  

\usepackage[pagebackref,breaklinks,colorlinks,citecolor=eccvblue]{hyperref}

\usepackage{orcidlink}

\begin{document}

\title{PolicyTrim: Boosting Intrinsic Policy Efficiency of Vision-Language-Action Models} 

\titlerunning{PolicyTrim}




\author{%
Xianghui Wang\inst{1}\thanks{Equal contribution.
$^\dagger$ Project lead.
$^{\ddagger}$Corresponding author.
},
Feng Chen\inst{2}$^{\star}$,
Wenbo Zhang\inst{2},
Hua Yan\inst{1},
Zixuan Wang\inst{1}$^\dagger$,
Changsheng Li\inst{3},
Yinjie Lei\inst{1}$^{\ddagger}$
}

\authorrunning{X.~Wang et al.}

\institute{%
{\small
$^{1}$Sichuan University
\quad
$^{2}$Adelaide University
\quad
$^{3}$Beijing Institute of Technology
}\\
{
\email{wangxianghui811421084@gmail.com}, \email{yinjie@scu.edu.cn}
}\\
}

\maketitle


\begin{abstract}
\emergencystretch=1em
Vision-Language-Action (VLA) models provide a unified paradigm for robotic manipulation, yet their real-world deployment is often bottlenecked by execution efficiency. While existing efforts predominantly focus on compute-centric efficiency to reduce per-step inference latency, the intrinsic \textbf{policy efficiency} of these models remains largely unexplored. Policy efficiency is fundamentally affected by two factors, namely the effective executable length of predicted action chunks and the total physical steps required to complete a task. These two factors jointly determine the total number of forward inference calls during execution. We observe that current VLA policies struggle with planning unreliability and action redundancy, suffering from severe prediction degradation at the tail of action chunks and tending to generate unnecessarily redundant physical steps. To address this, we propose \textbf{PolicyTrim}, a reinforcement learning-based post-training framework that extends the reliable action chunk length and reduces redundant physical steps. For reliable chunk extension, we employ a dynamic exploration strategy that explicitly rewards the successful completion of longer executable lengths, progressively pushing the trustworthy prediction horizon to its empirical limit. For step efficiency, we design a redundancy-aware reward that directly favors successful task completions with fewer steps while penalizing unreproducible shortcuts, effectively eliminating redundant physical actions. Extensive experiments across three benchmarks and three VLA models demonstrate that PolicyTrim improves action chunk utilization by 3$\times$ and reduces physical execution steps by 51.4\%. Ultimately, our framework delivers up to a 5.83$\times$ end-to-end deployment speedup without compromising task success rates.
\noindent\textbf{Project Page}: \url{https://inceptionwang.github.io/PolicyTrim/}

\end{abstract}

\begin{figure}[t]
  \centering
  \includegraphics[width=\textwidth]{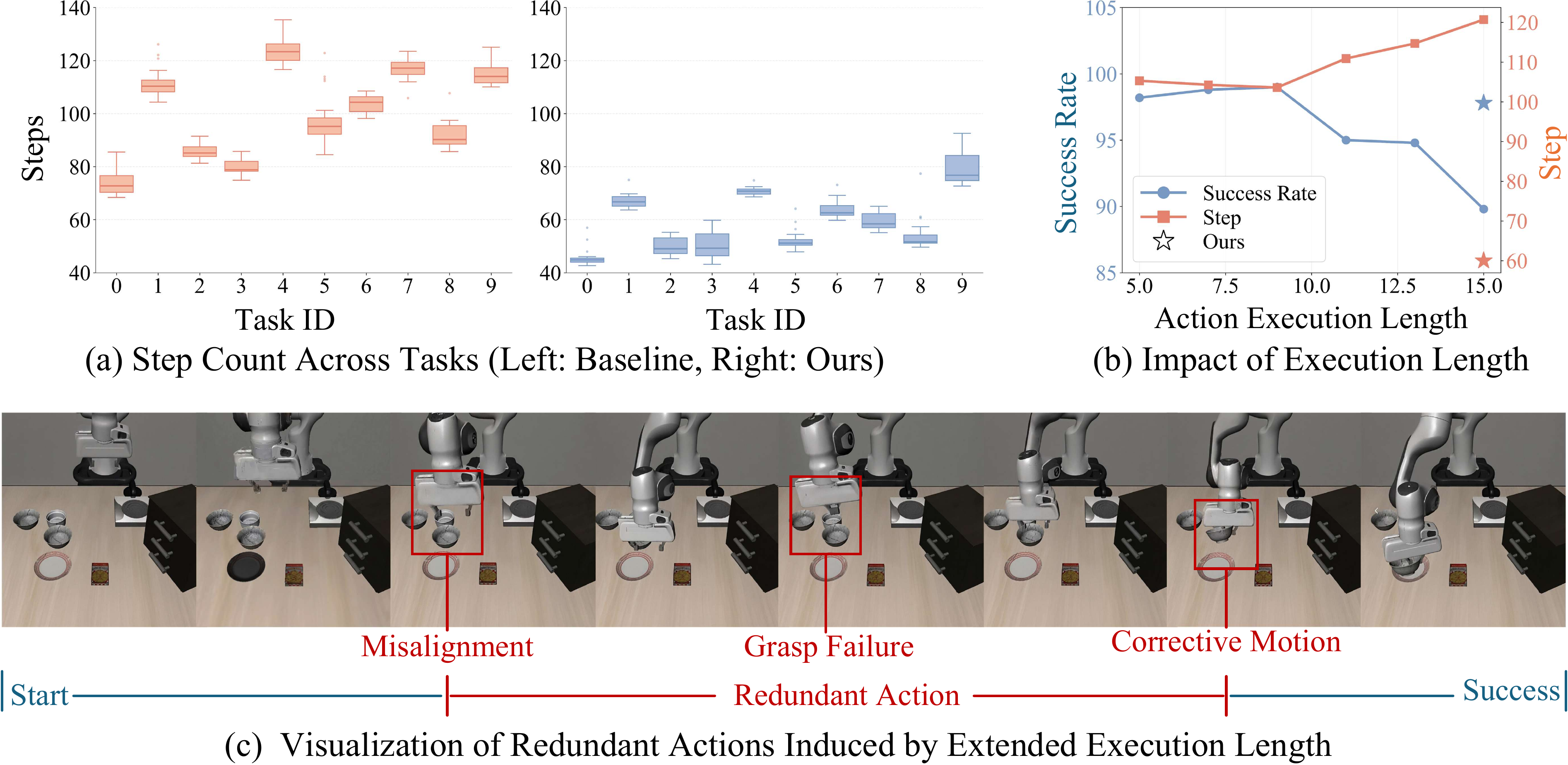}
  \caption{
  Intrinsic policy inefficiency in deployed VLA models manifests along two dimensions. (a) Repeated rollouts on identical tasks reveal substantial variance in step counts, indicating concise execution paths 
  exist but emerge only by chance. (b) Forcing longer action chunk execution simultaneously degrades success rates and inflates physical steps, confirming that unreliable tail predictions are a key factor. (c) A visualization of how tail prediction errors trigger misalignment and grasp failures, compelling the robot into redundant corrective actions before eventual task completion.
  }
  \label{fig:fig_1}
\end{figure}

\input{sec/intro}
\input{sec/related_work}
\input{sec/method}
\input{sec/conclusion}


%
%
\clearpage
\bibliographystyle{splncs04}
\bibliography{main}

\appendix
\input{sec/sup}

\end{document}

%% file: sec/intro.tex
\section{Introduction}
\emergencystretch=1em
Vision-Language-Action (VLA) models integrate visual perception, language understanding, and action generation into a single end-to-end framework, establishing a scalable paradigm for general-purpose robotic manipulation\cite{rt1_2022,palm_e_icml2023,rt2_corl2023,open_x_embodiment_rtx_2023,octo_rss2024,openvla_corl,pi0_rss2025,rdt1b_iclr2025}. 
To reduce the computational overhead of large vision-language backbones, the community has pursued compute-centric remedies such as visual token pruning \cite{li2025spvla}, quantization, and KV-cache reuse\cite{kvefficientvla_2025,retovla_2025,wen2025dvla,zheng2025ams,liu2025hybridvla}, all aimed at reducing per-inference latency.
However, the \textbf{policy efficiency} bottleneck of the models is largely unexplored, governed by the effective executable length of predicted action chunks and the total physical steps required to complete a task. These two factors jointly determine the total number of forward inference calls during execution.
Empirically, current VLA models face planning unreliability and action redundancy, exhibiting severe prediction degradation at the tail of action chunks and tending to generate redundant execution steps. As illustrated in Fig.~\ref{fig:fig_1}(b), when the model executes longer action chunks per inference, we observe a simultaneous decline in task success rates and an increase in the average physical steps required to complete the task. We believe this phenomenon stems from the degrading quality of tail predictions within action chunks~\cite{moh_2025,vla_knows_limits_2026}, resulting in an accumulation of physical errors that forces the robot to take redundant corrective actions. Furthermore, Fig.~\ref{fig:fig_1}(a) shows that rolling out a trained model multiple times on the same task reveals substantial variance in successful execution lengths. This indicates that more compact and efficient execution paths are physically reachable but currently emerge only by chance, leaving ample room for optimizing the model's step efficiency. Consequently, intrinsic policy efficiency remains the primary bottleneck for deployed VLA systems.

In this paper, we propose \textbf{PolicyTrim}, a two-stage RL-based post-training framework that enhances the policy efficiency of VLA models through reliable chunk extension and redundant step reduction\cite{deepseekmath_grpo_2024}. For reliable chunk extension, PolicyTrim diversifies the execution window lengths within a sampled group by assigning a fixed window size to each individual trajectory. This mechanism serves as a progressive reliability sweep to probe whether tail predictions at each chunk position remain trustworthy throughout actual task execution. Rollouts that successfully complete the task using longer action chunks receive higher rewards, progressively pushing the reliable planning frontier toward the empirical limit. Additionally, to achieve redundant step reduction, we design a step-saving reward that grants higher values to successful rollouts reaching the goal in fewer physical steps while a stability regularizer explicitly prevents the policy from collapsing into unreproducible shortcuts.
Through these two stages, PolicyTrim improves overall policy efficiency without requiring architectural modifications or additional expert data. The main contributions of this work are summarized as follows:
\begin{itemize}
    \item We identify policy efficiency as a critical yet overlooked deployment bottleneck for VLA models and distinguish it from pure computational efficiency by highlighting the dual challenges of unreliable tail predictions in action chunks and redundant physical execution steps.
    \item We propose PolicyTrim, a post-training framework that extends the reliable planning horizon and reduces step redundancy, without architectural changes or extra demonstrations.
    \item Experiments across multiple benchmarks and models show that PolicyTrim cuts physical execution steps in half, triples action chunk utilization, and delivers up to 5.83$\times$ end-to-end speedup without sacrificing success rate.
\end{itemize}

%% file: sec/related_work.tex
\section{Related Work}


\subsection{Vision-Language-Action Models}
Vision-Language-Action (VLA) models unify visual perception, language understanding, and action generation into a single end-to-end framework for robotic manipulation~\cite{roboflamingo_iclr2024,ma_survey_vla_embodied_2024,shao_survey_vlm_based_vla_2025,cogact_2024,gr1_2023}. Representative architectures span autoregressive VLAs such as OpenVLA~\cite{openvla_corl} and RT-2~\cite{rt2_corl2023}, as well as diffusion-based policies such as $\pi_0$~\cite{pi0_rss2025}, $\pi_{0.5}$~\cite{pi05_corl2025}, and GR00T~\cite{gr00t_n1_2025}. To handle high-frequency continuous control, modern VLAs widely adopt action chunking~\cite{act_rss2023,diffusion_policy_rss2023,openvla_oft_rss2025}, predicting a sequence of future actions at each decision step. However, the quality of predicted actions tends to degrade toward the tail of each chunk, and policies trained via imitation learning often exhibit unnecessarily redundant execution trajectories, both of which result in inefficient real-world deployment. While these models demonstrate strong task competence, their real-world deployment efficiency remains a critical concern. The end-to-end execution time of a VLA system is jointly determined by two factors: the per-step inference latency and the number of inference calls required to complete a task. The former is governed by computational efficiency, while the latter reflects the intrinsic policy efficiency of the deployed model. Despite growing attention to deployment efficiency, existing efforts have focused almost exclusively on the computational side, leaving policy efficiency largely unexplored.


\subsection{Efficient Vision-Language-Action Models}
Current efficiency methods target per-inference computational cost while treating the learned policy as fixed~\cite{yu_survey_efficient_vla_2025}. Visual token pruning~\cite{li2025spvla,wang2025specprunevla,jiang2025lightvla} and action tokenization compression~\cite{fast_rss2025,vqvla_iccv2025} reduce input and output overhead respectively. Speculative decoding~\cite{specvla_2025} and early-exit decoding~\cite{parallel_decoding_chunking_2025,ceedvla_2025} accelerate autoregressive generation, while KV-cache~\cite{kvefficientvla_2025,retovla_2025,xu2025vlacache} reuse further reduces decoding latency. Model quantization~\cite{bitvla_2025,quantvla_2026,qvla_2026,sqapvla_2025} and lightweight designs~\cite{smolvla_2025,edgevla_2025,wen2025tinyvla,hung2025nora} further lower hardware demands. By strictly preserving the original policy distribution, these methods inevitably inherit its intrinsic inefficiencies~\cite{efficientvla_2025}, such as unreliable tail predictions within action chunks and redundant physical execution steps, which remain unaddressed regardless of how aggressively inference is accelerated. Critically, when a policy requires an excessive number of inference calls to complete a task, gains from per-step acceleration are fundamentally bounded, as the total execution time is jointly determined by both latency and call frequency. In contrast, improving policy efficiency directly reduces the number of inference calls required, offering a complementary and multiplicative source of speedup. Since the two axes are fully decoupled, advances in either dimension can be stacked to yield compounded gains beyond what either approach achieves independently. Policy efficiency therefore constitutes an orthogonal axis to computational efficiency, one that requires post-training intervention rather than architectural or hardware-level optimization alone.


\subsection{Reinforcement Learning for VLA}
Reinforcement learning has emerged as a post-training paradigm to push VLA policies beyond their initial capabilities~\cite{vla_rl_2025,co_rft_2025,simplevla_rl_2025,conrft_rss2025}. Early efforts applied PPO~\cite{ppo_2017}, but the actor-critic architecture introduces prohibitive memory overhead when scaling to large VLA backbones~\cite{lora_2021}. GRPO~\cite{deepseekmath_grpo_2024} eliminates the value model by computing group-relative advantages, making RL post-training practical for large-scale VLAs. However, existing GRPO approaches for VLAs universally rely on binary success rewards~\cite{vla_rl_2025,co_rft_2025,simplevla_rl_2025,chen2025sparsity}, which create two fundamental limitations. First, once the policy achieves a high success rate, reward variance within the sampled group collapses, causing advantage estimates to lose discriminative power and learning to stagnate. Second, binary rewards provide no signal to distinguish shorter completions from longer ones, nor to incentivize extending the reliable prediction horizon beyond the conservative default, leaving execution efficiency entirely unoptimized. PolicyTrim addresses both limitations through a step-saving reward that maintains meaningful optimization signal even at high success rates and a progressive chunk exploration mechanism that actively pushes the trustworthy prediction frontier toward the architectural limit.

%% file: sec/method.tex
\section{Method}
\emergencystretch=1em

\subsection{Overview}
\label{sec:overview}

\begin{figure}[!t]
  \centering
  \includegraphics[width=\textwidth]{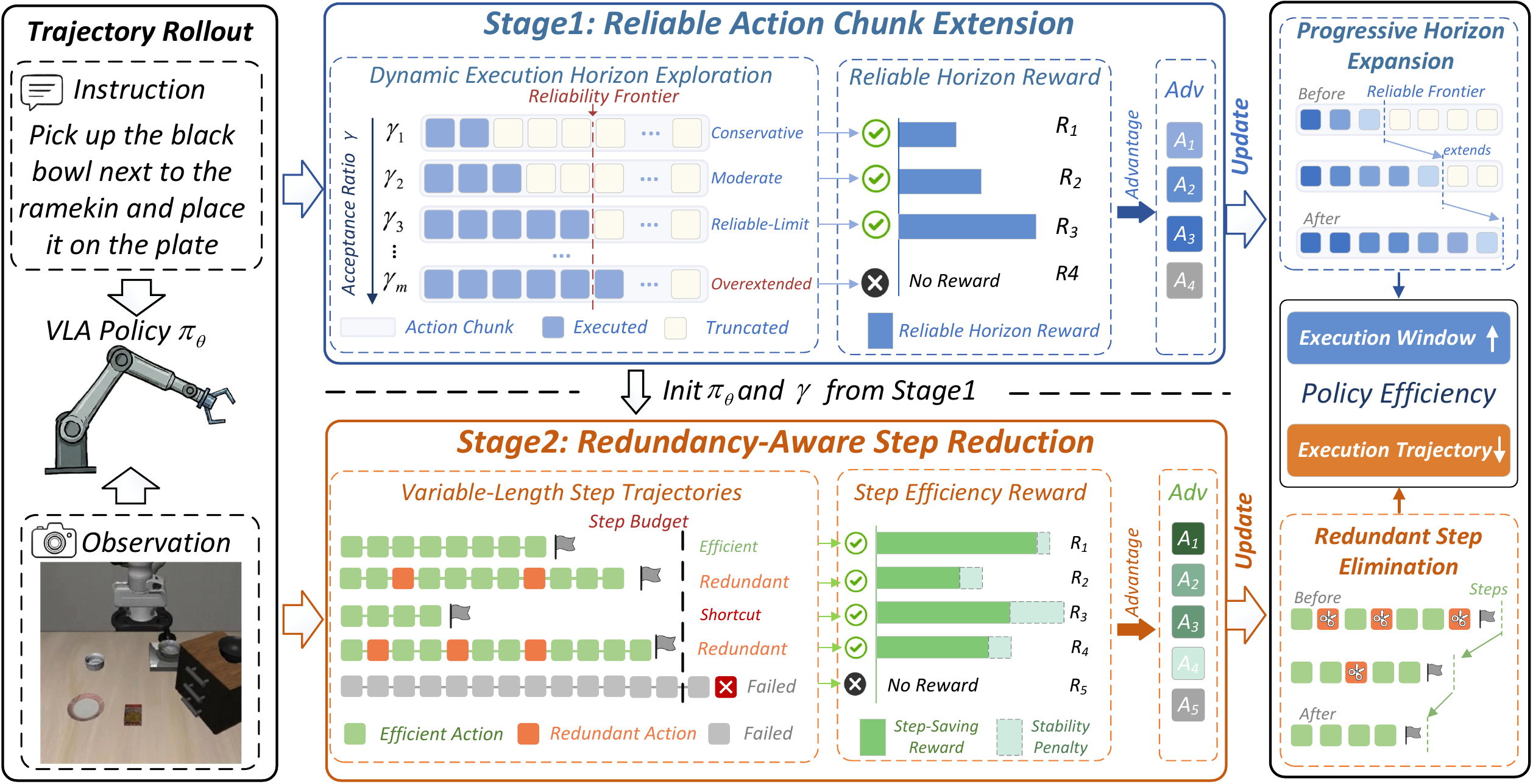}
\caption{
\textbf{Overview of PolicyTrim.} PolicyTrim is a two-stage RL post-training framework that enhances intrinsic policy efficiency of VLA models. The first stage progressively extends the reliable action chunk horizon by rewarding successful execution of longer chunks. The second stage eliminates redundant physical steps via a step-saving reward coupled with group-anchored stability regularization. Together, the two stages jointly reduce the total number of forward inference calls required to complete a task.
}
  \label{fig:fig_2}
\end{figure}

An overview of PolicyTrim is illustrated in Fig.~\ref{fig:fig_2}. We propose a two-stage post-training framework that extends the executable action horizon per inference and reduces the number of steps required to complete a task for VLA models.
At an arbitrary decision step $t$, the policy $\pi_\theta$ processes the current visual observation $o_t$ and language instruction $l$ to predict a sequence of future actions $a_{t:t+H}$ in parallel, where $H$ denotes the maximum chunk capacity. Since prediction quality degrades toward the tail of action chunks, the system is typically constrained to executing only a truncated subset of actions to avoid the accumulation of physical errors. This subset defines an execution window $h = \lfloor \gamma H \rfloor$ before the model acquires a new observation for the next planning cycle, where $\gamma \in (0, 1]$ represents the acceptance ratio. While this conservative execution strategy helps maintain physical stability, it handicaps policy efficiency by discarding potentially reliable predictions and necessitating frequent inference calls.

This paper introduces \textbf{PolicyTrim}, an RL-based post-training framework designed to enhance policy efficiency. As illustrated in Fig.~\ref{fig:fig_2}, the framework decouples this enhancement objective into two progressive learning stages targeting reliable action chunk extension and redundancy-aware step reduction respectively. To facilitate reliable action chunk extension, the initial stage diversifies the execution lengths within a sampled group by assigning a different acceptance ratio to each individual trajectory. This mechanism implements a progressive reliability sweep that probes the empirical boundaries of trustworthy predictions. By rewarding successful trajectories proportionally to their chunk utilization, the model learns to yield more reliable action predictions further into the chunk tail and progressively extends the execution window. Building upon these extended chunks, a complementary redundancy-aware mechanism subsequently drives the second stage of physical step reduction.
 This phase introduces a step-saving reward coupled with a per-trajectory stability penalty. This method explicitly incentivizes the policy to prune redundant intermediate actions and converge toward the most concise physical execution trajectory while preventing catastrophic collapse onto irreproducible shortcuts. Ultimately, this sequential refinement of expanding reliable action chunks and reducing total steps naturally leads to a significant reduction in the required inference calls.

\subsection{Reliable Action Chunk Extension}
\label{subsec:reliable_chunk}
To enable reliable predictions over extended action horizons, the proposed method avoids rigidly forcing the model to execute the maximum length from scratch. It is instead framed as a progressive extension process where the framework dynamically probes the empirical reliability of various chunk positions by assigning execution windows of varying lengths across sampled trajectories. The policy is explicitly incentivized to overcome the inherent tail degradation by assigning higher rewards to those successful rollouts that manage to sustain longer execution length. Ultimately, this progressive refinement paradigm seamlessly pushes the trustworthy prediction frontier toward the maximum usable action chunk length $H$ supported by the underlying model architecture.

\noindent\textbf{Dynamic Execution Horizon Exploration.}
To probe the empirical reliability boundary without additional rollouts, each trajectory within the group is assigned a distinct execution window, collectively spanning multiple prediction horizons within a single group. Formally, we define a discrete set of acceptance ratios $\Gamma=\{\gamma_{1},\gamma_{2},\ldots,\gamma_{M}\}$ with $0<\gamma_{m}\le1$, and assign each trajectory $\tau_{i}$ its own ratio $\gamma_{i}$, yielding an execution window $h_{i}=\lfloor\gamma_{i}H\rfloor$. This per-trajectory assignment turns each sampled group into a reliability sweep, probing prediction trustworthiness across short, intermediate, and long chunk positions in parallel.

\noindent\textbf{Reliable Horizon Reward.}
The reward for chunk extension is composed of a task completion reward 
and a horizon reward. The task completion reward $R_{succ}(\tau_i)$ assigns 
$1$ to a successful trajectory and $0$ otherwise. The horizon reward is 
scaled by the assigned acceptance ratio:
\begin{equation}
    R_{horizon}(\tau_i)=\beta\cdot\gamma_i,
\end{equation}
where a larger $\gamma_i$ indicates that the model sustained accurate 
predictions over a longer execution window without intermediate 
re-observation. However, a naive combination of the two becomes 
horizon-biased when no trajectory in the group succeeds, because 
the task completion signal collapses to a constant while the horizon 
reward still pushes the policy toward longer but potentially erroneous 
execution. To incentivize horizon extension without collapsing to 
unreliable long-chunk behaviors, we activate the horizon reward only 
when the group contains at least one successful trajectory. Concretely, 
the integrated reward for chunk extension is formulated as:
\begin{equation}
    R_{ext}(\tau_i)=\mathcal{I}_{succ}^{(i)}\cdot(R_{succ}(\tau_i)+R_{horizon}(\tau_i)),
\end{equation}
where $\mathcal{I}_{succ}^{(i)} \in \{0, 1\}$ is a binary indicator 
that equals $1$ only if trajectory $\tau_i$ successfully completes the task.

\noindent\textbf{Group-Relative Policy Update.} 
We optimize the aforementioned objective utilizing Group Relative Policy Optimization (GRPO). For a group of $G$ trajectories sampled from the current policy $\pi_{\theta}$ under the identical initial state $o_t$ and language instruction $l$, we compute a group-normalized advantage for each trajectory based on the current objective $R_{ext}(\tau_i)$:
\begin{equation}
    A_{i}=\frac{R_{ext}(\tau_i)-\mu_{R}}{\sigma_{R}+\epsilon},
    \label{eq:advantage}
\end{equation}
where $\mu_{R}$ and $\sigma_{R}$ are the mean and standard deviation of $R_{ext}(\tau_i)$ within the group. The advantage computation then directly contrasts these varying execution lengths under the same task instance, enabling the policy to learn which chunk positions still yield trustworthy actions and where prediction quality begins to degrade. Subsequently, the policy is updated via a clipped surrogate objective with a Kullback-Leibler (KL) penalty to the reference policy $\pi_{ref}$:
\begin{equation}
    L(\theta)=\mathbb{E}_{i}\left[\min\left(r_{i}(\theta)A_{i}, \text{clip}(r_{i}(\theta),1-\epsilon,1+\epsilon)A_{i}\right)\right]-\beta_{KL}D_{KL}(\pi_{\theta}||\pi_{ref}),
    \label{eq:policy_update}
\end{equation}
where $r_{i}(\theta)$ denotes the importance sampling ratio between the updated policy and the old policy distribution, and $\epsilon$ defines the clipping range that constrains $r_{i}(\theta)$ within $(1-\epsilon, 1+\epsilon)$ to prevent excessively large policy updates. The KL divergence $D_{KL}(\pi_{\theta}||\pi_{ref})$ is computed per-token across the generated action chunk to prevent substantial deviation from the manipulation priors established during pre-training.

\subsection{Redundancy-Aware Step Reduction}
\label{subsec:step_efficient}
To explicitly encourage concise execution while preserving task correctness, we introduce a step-saving reward for successful trajectories, complemented by a group-consistent step regularization to prevent exploitation of fragile shortcuts.

\noindent\textbf{Step-Saving Reward.}
 We define a step budget $S_{base}$ based on the initial policy's performance statistics. For a successful trajectory completing the task in $S(\tau_i)$ steps, the step-saving reward is defined as:
\begin{equation}
    R_{step}(\tau_i)=\frac{\max(0,S_{base}-S(\tau_i))}{S_{base}}.
\end{equation}
Trajectories that complete the task in fewer steps receive proportionally higher reward, directly incentivizing the policy to converge toward more concise execution trajectories.


\noindent\textbf{Group-Anchored Regularization.} Naively minimizing execution steps without constraints risks policy collapse. Due to high initial variance in step counts, occasional short trajectories may receive disproportionately large rewards, causing the policy to exploit shortcuts that are not reliably reproducible. To address this, we introduce a group-anchored stability penalty that regularizes the step distribution within each sampled group:

\begin{equation}
    P_{stab}(\tau_i)=\lambda_{stab}\cdot\tanh\!\left(
    \frac{|S(\tau_i)-\mu_{group}|}{\max(\sigma_{group},\,\sigma_{floor})}
    \right),
\end{equation}
where $\mu_{group}$ and $\sigma_{group}$ are the mean and standard deviation of $S(\tau)$ computed over successful trajectories in the group, serving as a group consensus anchor. The $\tanh$ function smoothly maps the normalized deviation to a bounded penalty in $[0, 1]$, ensuring that trajectories deviating substantially from the group consensus incur progressively larger penalties while avoiding unbounded penalization. This design discourages the policy from exploiting fragile step-count outliers, guiding it to shift smoothly toward higher efficiency rather than collapsing onto irreproducible shortcuts. 
Meanwhile, as the policy converges to a consistent execution regime, $\sigma_{group}$ may become very small, causing even minor step deviations to incur disproportionately large penalties. To address this, we introduce a floor parameter $\sigma_{floor}$ to bound the denominator, preventing such over-penalization and ensuring that subsequent policy updates remain well-conditioned once a stable regime is reached.


\noindent\textbf{Joint Step-Performance Reward.} The final reward for step reduction integrates the task completion reward $R_{succ}$, the step-saving reward $R_{step}$, and the stability penalty $P_{stab}$:
\begin{equation}
    R_{eff}(\tau_i)=\mathcal{I}_{succ}^{(i)}\cdot(R_{succ}(\tau_i)+R_{step}(\tau_i)-P_{stab}(\tau_i)),
\end{equation}
where the binary indicator $\mathcal{I}_{succ}^{(i)}$ ensures that failed rollouts strictly receive a reward of zero, preventing step-saving incentives from reinforcing unsuccessful behaviors. We then utilize these rewards to calculate the group-relative advantage using Eq.~(\ref{eq:advantage}) and execute the final policy update following Eq.~(\ref{eq:policy_update}). 

Overall, inspired by the empirical insights from Fig.~\ref{fig:fig_1}, we decouple the optimization into two sequential stages. Reliable chunk extension first establishes a wider trustworthy prediction horizon. However, committing to longer action chunks alone provides no mechanism to constrain the total number of physical steps, and may even inflate it through compounded tail prediction errors. Conversely, jointly optimizing both objectives simultaneously conflates two fundamentally distinct sub-goals, potentially entangling the reward signal and complicating the optimization landscape. We therefore apply redundancy-aware step reduction as a subsequent stage, building upon the extended chunk horizon to compress the physical execution path and systematically address the dual bottlenecks of VLA models.
\label{sec:experiments}
\emergencystretch=1em

\section{Experiment}
\subsection{Experimental Setup}

\paragraph{Benchmarks.} We evaluate on three diverse benchmarks including LIBERO~\cite{libero_neurips2023_db}, ManiSkill~\cite{taomaniskill3}, Meta-World~\cite{mclean2025metaworld} and further validate its
sim-to-real transfer on a physical robot platform. Reported metrics
include average success rate, average physical steps, average action
chunk execution length, end-to-end execution speedup, and wall-clock
execution time for real-world deployment.
\begin{itemize}
    \item[$\bullet$] \textbf{LIBERO} is a tabletop manipulation benchmark comprising four subsets of increasing difficulty. The Spatial and Object subsets evaluate spatial reasoning and object generalization, the Goal subset introduces goal-conditioned reasoning, and the Long subset requires multi-stage manipulation over extended horizons, making it well-suited for assessing step efficiency under prolonged execution.
    \item[$\bullet$] \textbf{ManiSkill} is a high-fidelity simulation platform offering physics-rich continuous control tasks. Following the experimental setup in~\cite{yu2025rlinf}, we adopt its diverse pick-and-place task combinations to evaluate policy efficiency and cross-task generalization in precise manipulation scenarios.
    \item[$\bullet$] \textbf{Meta-World} covers a wide range of manipulation tasks beyond pick-and-place, encompassing diverse end-effector motions and interaction modes. We adopt the MT50 suite to assess policy efficiency across heterogeneous action spaces and manipulation dynamics.
    \item[$\bullet$] \textbf{Real-world Deployment} uses an Agilex Piper arm
    equipped with two Intel RealSense D435i cameras. We evaluate three
    tabletop manipulation tasks, FlipMug, HangMug, and TapeBox.

\end{itemize}

\paragraph{VLA Models.} To verify cross-architecture generalization, we apply PolicyTrim to three VLA models spanning distinct architectural paradigms. $\pi_{0.5}$ is a conditional diffusion policy that generates action chunks through iterative flow-matching denoising, conditioned on vision-language features from a pretrained VLM backbone. Its stochastic decoding naturally captures complex action distributions. OpenVLA-OFT builds upon a 7B-parameter vision-language model and replaces autoregressive token generation with parallel chunk decoding via placeholder action tokens and bidirectional attention, enabling single-forward-pass prediction of entire action sequences. GR00T is a generalist robot transformer that adopts a dual-system design, pairing a slow vision-language reasoning module with a fast diffusion-based action generation head.

\paragraph{Implementation Details.} All experiments are built upon the RLinf framework~\cite{yu2025rlinf}. We use a group size of $G = 8$ trajectories for each task in every iteration. For all VLA models, the maximum chunk capacity $H$ predicted at each step is initialized to match or exceed the original settings of the respective checkpoints. For reliable chunk extension, the sampling set of the acceptance ratio is $\Gamma = \{\gamma_1, \gamma_2, \dots, \gamma_M \}$ with $0 < \gamma_m \le 1$ and $M = 3$ by default; $\gamma_1$ is set such that $\gamma_1 \cdot H$ equals the model's original default execution length, and the remaining ratios are uniformly spaced up to $1$. For step efficiency, the step budget $S_{base}$ is set to approximately 1.3 times the average successful steps of the initial baseline policy. All experiments are conducted on 8 A100 GPUs with 64 parallel simulation environments.

\begin{table}[!t]
\centering
\caption{Evaluation of $\pi_{0.5}$, OpenVLA-OFT, and GR00T on the four subsets of the LIBERO benchmark. We report average success rate (SR), average physical steps ($S_{\text{total}}$), average action chunk execution length ($h_{\text{chunk}}$), and end-to-end execution Speedup (Spd).}
\label{tab:libero_all_models}
\resizebox{\textwidth}{!}{
\begin{tabular}{ll cccc cccc cccc}
\toprule
\multirow{2}{*}{Task} & \multirow{2}{*}{Method} & \multicolumn{4}{c}{$\pi_{0.5}$} & \multicolumn{4}{c}{OpenVLA-OFT} & \multicolumn{4}{c}{GR00T} \\
\cmidrule(lr){3-6} \cmidrule(lr){7-10} \cmidrule(lr){11-14}
& & SR & $S_{\text{total}}$ & $h_{\text{chunk}}$ & Spd$\uparrow$ & SR & $S_{\text{total}}$ & $h_{\text{chunk}}$ & Spd$\uparrow$ & SR & $S_{\text{total}}$ & $h_{\text{chunk}}$ & Spd$\uparrow$ \\
\midrule
\multirow{2}{*}{Spatial}
& Baseline   & 97.8 & 108.3 & 5 & 1.0 & 98.6 & 111.2 & 8 & 1.0 & 91.4 & 67.2 & 5 & 1.0 \\
& PolicyTrim & 97.8 & 59.8 & 15 & 5.43 & 98.8 & 62.1 & 8 & 1.79 & 92.0 & 56.6 & 10 & 2.37 \\
\midrule
\multirow{2}{*}{Object}
& Baseline   & 99.1 & 125.0 & 5 & 1.0 & 98.5 & 135.2 & 8 & 1.0 & 95.0 & 71.3 & 5 & 1.0 \\
& PolicyTrim & 98.5 & 64.3 & 15 & 5.83 &  98.5 & 68.8 & 8 & 1.97 & 95.3 & 65.5 & 10 & 2.18 \\
\midrule
\multirow{2}{*}{Goal}
& Baseline   & 98.7 & 110.6 & 5 & 1.0 & 97.7 & 118.6 & 8 & 1.0 & 84.2 & 63.3 & 5 & 1.0 \\
& PolicyTrim & 98.8 & 63.5 & 15 & 5.23 & 98.0 & 66.9 & 8 & 1.77 & 86.3 & 60.8 & 10 & 2.08 \\
\midrule
\multirow{2}{*}{Long}
& Baseline   & 93.0 & 249.8 & 5 & 1.0 & 92.9 &249.3 & 8 & 1.0 & 86.1 & 177.9 & 5 & 1.0 \\
& PolicyTrim & 93.3 & 171.8 & 10 & 2.91 & 93.1 & 178.3 & 8 & 1.40 & 89.2 & 165.9 & 10 & 2.14 \\
\bottomrule
\end{tabular}
}
\end{table}

\begin{table}[!t]
\centering
\caption{Evaluation on ManiSkill and Meta-World. We report average success rate (SR), average physical steps ($S_{\text{total}}$), average action chunk execution length ($h_{\text{chunk}}$), and end-to-end execution Speedup (Spd).}
\label{tab:maniskill_metaworld}
\begin{tabular}{cl cccc cccc}
\toprule
\multirow{2}{*}{Benchmark} & \multirow{2}{*}{Method} & \multicolumn{4}{c}{$\pi_{0.5}$} & \multicolumn{4}{c}{OpenVLA-OFT} \\
\cmidrule(lr){3-6} \cmidrule(lr){7-10}
& & SR & $S_{\text{total}}$ & $h_{\text{chunk}}$ & Spd$\uparrow$ & SR & $S_{\text{total}}$ & $h_{\text{chunk}}$ & Spd$\uparrow$ \\
\midrule
\multirow{2}{*}{ManiSkill}
& Baseline     & 88.1 & 45.2 & 5 & 1.0         & 60.6 & 53.1 & 8 & 1.0 \\
& PolicyTrim   & 89.8 & 38.3 & 10 & 2.36         & 63.2 & 46.7 & 8 & 1.14 \\
\midrule
\multirow{2}{*}{Meta-World}
& Baseline     & 65.1 & 66.3 & 5 & 1.0 & \multicolumn{4}{c}{--} \\
& PolicyTrim   & 65.4 & 52.6 & 10 & 2.52         & \multicolumn{4}{c}{--} \\
\bottomrule
\end{tabular}
\end{table}

\begin{figure}[!t]
  \centering
  \includegraphics[width=\textwidth]{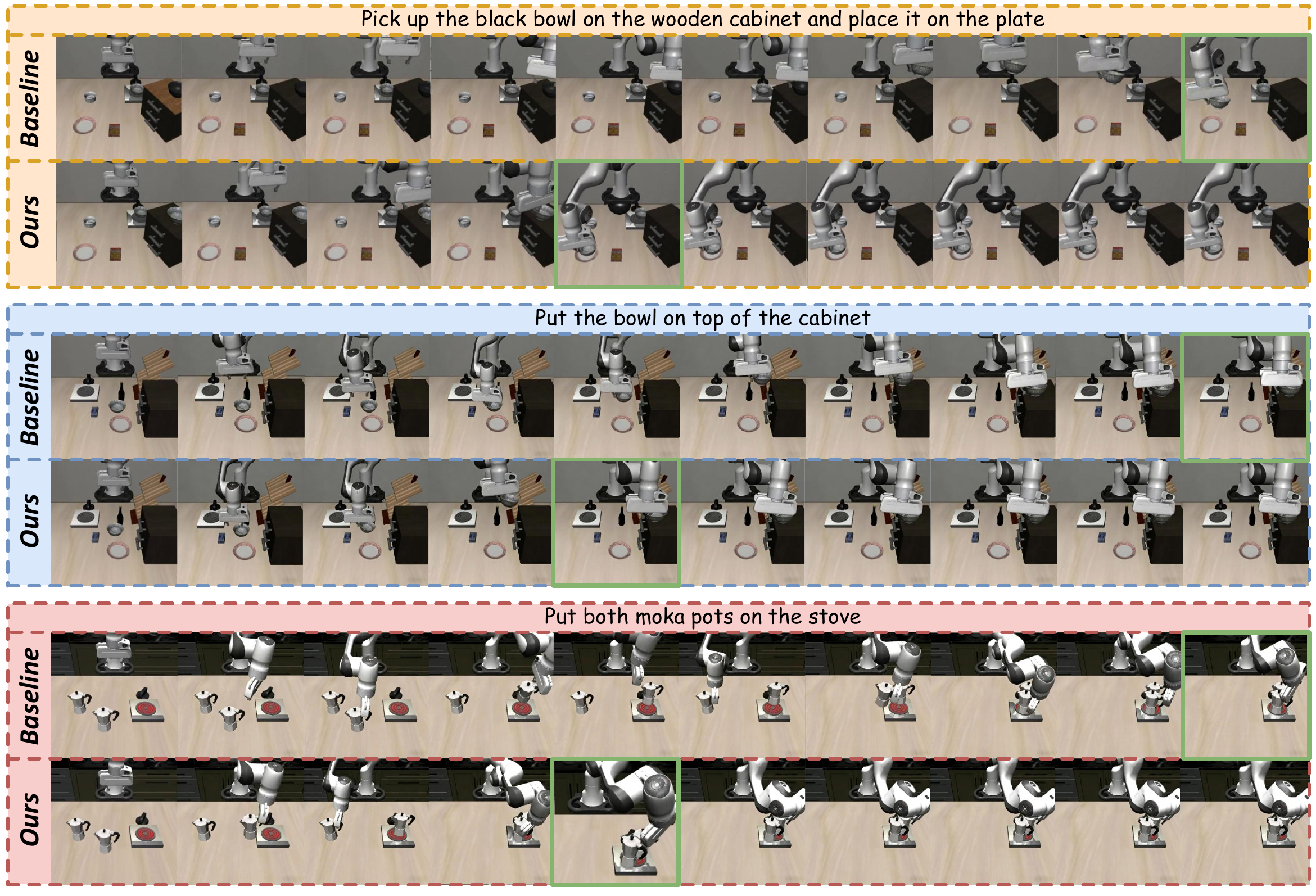}
  \caption{Qualitative comparison on randomly sampled LIBERO tasks. Under identical configurations, the baseline incurs redundant physical actions, whereas PolicyTrim achieves task completion in roughly half the steps.}
  \label{fig:qualitative}
\end{figure}

\subsection{Main Results}

\paragraph{Results on LIBERO.} Table~\ref{tab:libero_all_models} presents a comprehensive quantitative evaluation on the four LIBERO subsets using $\pi_{0.5}$, OpenVLA-OFT, and GR00T. Across all three models and subsets, PolicyTrim consistently reduces physical execution steps while maintaining comparable task success rates, with the magnitude of reduction varying by architecture and task complexity. The most pronounced reduction is observed for $\pi_{0.5}$ on the Object subset, where the step count drops to as low as 51.4\% of the baseline. These results suggest that our efficiency gains arise from a fundamentally improved execution strategy rather than any sacrifice in task competence.
For action chunk extension, we observe an architectural divergence among the evaluated models. $\pi_{0.5}$ and GR00T both adopt diffusion-based action decoding, which naturally accommodates extended chunk horizons; PolicyTrim successfully increases the reliable action chunk execution length for both architectures. OpenVLA-OFT, by contrast, employs a parallel decoding scheme with placeholder action tokens and bidirectional attention, where even a marginal increase in chunk length causes severe accuracy degradation, rendering chunk extension effectively untrainable for this architecture. We therefore apply only the step reduction stage to OpenVLA-OFT, and the reported chunk length improvements are specific to $\pi_{0.5}$ and GR00T.


\paragraph{Results on ManiSkill and Meta-World.} Table~\ref{tab:maniskill_metaworld} reports the performance on ManiSkill and Meta-World. PolicyTrim consistently improves both success rates and step efficiency across benchmarks, achieving up to 2.52$\times$ end-to-end speedup on Meta-World and 2.36$\times$ on ManiSkill with $\pi_{0.5}$. The consistent improvements across diverse physical simulators, action spaces, and model architectures demonstrate that PolicyTrim generalizes well beyond the LIBERO benchmark, confirming that policy efficiency optimization yields substantial and reliable gains across varied deployment scenarios.

\paragraph{Architectural Generality.}
Table~\ref{tab:architecture_main} reports the cross-architecture results on both parallel-decoding and autoregressive VLA models. Beyond the standard OpenVLA-OFT setting in Table~\ref{tab:libero_all_models}, we re-pretrain OpenVLA-OFT with a larger action chunk capacity of $h=16$, since the original parallel decoder uses a fixed horizon of $h=8$ and already executes all predicted actions, leaving no room for Stage~1 extension. With this re-pretrained OpenVLA-OFT backbone, the full two-stage PolicyTrim pipeline achieves a $2.97\times$ speedup while preserving task success. We further evaluate the autoregressive OpenVLA model, where Stage~1 is not directly applicable because action chunks are not generated through the same fixed-horizon parallel decoding mechanism. Applying Stage~2 alone still yields a $1.41\times$ speedup and improves the success rate from $84.7\%$ to $87.0\%$. These results demonstrate that PolicyTrim provides consistent gains across different VLA decoding architectures, and that its redundancy-aware optimization is not tied to a specific action decoding form.

\begin{table}[!t]
\centering
\caption{Cross-architecture results. We report success rate (SR), average physical steps, action horizon $h$, and end-to-end speedup.}
\label{tab:architecture_main}
\setlength{\tabcolsep}{5pt}
\begin{tabular}{llcccc}
\toprule
Model & Method & SR & Step & $h$ & Spd$\uparrow$ \\
\midrule
OpenVLA-OFT & Baseline & 98.6 & 111.2 & 8  & 1.00$\times$ \\
OpenVLA-OFT & S1+S2   & 98.8 &  65.4 & 14 & 2.97$\times$ \\
\midrule
OpenVLA & Baseline & 84.7 & 113.5 & -- & 1.00$\times$ \\
OpenVLA & S2       & 87.0 &  80.6 & -- & 1.41$\times$ \\
\bottomrule
\end{tabular}
\end{table}

\paragraph{Real-World Deployment.}
Table~\ref{tab:real_world} reports the real-world deployment results of $\pi_{0.5}$ on three tabletop manipulation tasks, FlipMug, HangMug, and TapeBox. PolicyTrim maintains or improves the success rate across all tasks, under both the standard setting with a fixed target pose and the dynamic setting where the target is randomly perturbed during grasping, while substantially reducing wall-clock execution time. On average, PolicyTrim achieves a $1.86\times$ speedup over the baseline under the standard real-world setting. These results are obtained on an Agilex Piper arm equipped with two Intel RealSense D435i cameras, where PolicyTrim is first RL post-trained in simulation and then adapted to the real world through supervised fine-tuning on a small number of real demonstrations. The consistent reduction in real execution time demonstrates that PolicyTrim's efficiency gains transfer from simulation to physical deployment.

\begin{table}[!t]
\centering
\caption{Real-world deployment results. Standard uses a fixed target pose, while Dynamic perturbs the target during grasping. Values under Standard and Dynamic are success rates in \%, and Time is measured in seconds.}
\label{tab:real_world}
\begin{tabular}{lccccc|ccc}
\toprule
& \multicolumn{3}{c}{Standard}
& \multicolumn{2}{c|}{Dynamic}
& \multicolumn{3}{c}{Time(Standard)} \\
\cmidrule(lr){2-4} \cmidrule(lr){5-6} \cmidrule(lr){7-9}
Method & Flip & Hang & Tape & Flip & Tape & Flip & Hang & Tape \\
\midrule
Baseline   & 70 & 60 & 95 & 70 & 65 & 14.6 & 15.6 & 17.5 \\
PolicyTrim & 75 & 65 & 95 & 70 & 70 &  7.6 &  8.7 &  9.4 \\
\bottomrule
\end{tabular}
\end{table}

\paragraph{Qualitative Analysis.} Fig.~\ref{fig:qualitative} provides a qualitative comparison between the baseline and PolicyTrim on tasks randomly sampled from LIBERO under identical configurations. The baseline consistently exhibits redundant corrective motions before task completion, whereas PolicyTrim executes smooth and direct trajectories toward the goal. This behavioral contrast is especially pronounced in tasks involving precise placement, where the baseline displays noticeable hesitation and jittering near the target, substantially inflating the total step count. PolicyTrim avoids such corrective detours entirely, reducing physical steps by nearly half. 

\begin{table}[!t]
\centering
\caption{Ablation study of different components on LIBERO-Spatial benchmarks.}
    \setlength{\tabcolsep}{4pt} 
    \begin{tabular}{ccccccc}
    \toprule
    Reliable Chunk & Step-Saving & Group-Anchored & \multirow{2}{*}{SR} & \multirow{2}{*}{$S_{\text{total}}$} & \multirow{2}{*}{$h_{\text{chunk}}$} & \multirow{2}{*}{Spd$\uparrow$} \\
    Extension & Reward & Regularization & & & & \\
    \midrule
    \multicolumn{1}{c}{\textcolor[gray]{0.8}{\ding{55}}} & \multicolumn{1}{c}{\textcolor[gray]{0.8}{\ding{55}}} & \multicolumn{1}{c}{\textcolor[gray]{0.8}{\ding{55}}} & 97.8 & 108.3 & 5 & 1.0 \\
    \multicolumn{1}{c}{\ding{51}} & \multicolumn{1}{c}{\textcolor[gray]{0.8}{\ding{55}}} & \multicolumn{1}{c}{\textcolor[gray]{0.8}{\ding{55}}} & 97.2 & 113.8 & 15 & 2.86 \\
    \multicolumn{1}{c}{\textcolor[gray]{0.8}{\ding{55}}} & \multicolumn{1}{c}{\ding{51}} & \multicolumn{1}{c}{\textcolor[gray]{0.8}{\ding{55}}} & 93.7 & 81.7 & 5 & 1.32 \\
    \multicolumn{1}{c}{\textcolor[gray]{0.8}{\ding{55}}} & \multicolumn{1}{c}{\ding{51}} & \multicolumn{1}{c}{\ding{51}} & 97.5 & 61.6 & 5 & 1.75 \\
    \multicolumn{1}{c}{\ding{51}} & \multicolumn{1}{c}{\ding{51}} & \multicolumn{1}{c}{\ding{51}} & 98.3 & 59.8 & 15 & 5.43 \\
    \bottomrule
    \end{tabular}
\label{tab:ablation_conmpo}
\end{table}

\begin{table}[!t]
\centering
\caption{Ablation of Dynamic Execution Horizon Exploration on LIBERO-Object 
using $\pi_{0.5}$ with $H\!=\!20$. Fixed-$\gamma$ variants replace 
diverse ratio sampling with a single acceptance ratio.}
\label{tab:ablation_object}
\setlength{\tabcolsep}{4pt}
\begin{tabular}{lcccc}
\toprule
Method & SR & $S_{\text{total}}$ & $h_{\text{chunk}}$ & Spd$\uparrow$ \\
\midrule
Baseline                    & 99.1 & 125.0 &  5 & 1.00 \\
\midrule
Fixed $\gamma\!=\! 0.25$     & 99.2 & 127.1 &  5 & 0.98 \\
Fixed $\gamma\!=\! 0.50$     & 98.8 & 125.2 & 10 & 1.99 \\
Fixed $\gamma\!=\! 0.75$     & 95.8 & 130.1 & 15 & 2.88 \\
Fixed $\gamma\!=\! 1.0$      & 94.4 & 131.8 & 20 & 3.79 \\
\midrule
Dynamic Horizon Exploration & 98.8 & 127.4 & 15 & 2.94 \\
\bottomrule
\end{tabular}
\end{table}

\subsection{Ablation Study}
\emergencystretch=1em
In Table~\ref{tab:ablation_conmpo}, we conduct ablation studies on the LIBERO-Spatial subset using the $\pi_{0.5}$ model to isolate the contribution of each PolicyTrim component. All ablated configurations share the same initialization checkpoint and hyperparameters, with only the target component removed.

\paragraph{Effect of Reliable Action Chunk Extension.}
Adding Chunk Extension alone triples the average execution window 
from $h_{\text{chunk}}\!=\!5$ to $15$, reducing inference call 
frequency and yielding a $2.86\times$ speedup. However, total 
physical steps simultaneously increase from $108.3$ to $113.8$, 
confirming that committing to longer action chunks amplifies the 
downstream impact of residual tail prediction errors and compels 
the robot to take additional corrective actions. This validates 
the necessity of an explicit step reduction stage following chunk 
extension, as longer chunks alone do not translate to more 
concise physical execution.

\paragraph{Effect of Step-Saving Reward.} Applying the Step-Saving Efficiency Reward alone successfully reduces $S_{\text{total}}$ by $24.6\%$ from $108.3$ to $81.7$, demonstrating its effectiveness in driving the policy toward more concise execution. However, this comes at the cost of an unacceptable degradation in task competence, with the success rate dropping sharply from $97.8\%$ to $93.7\%$. Without Group-Anchored Regularization, the policy exploits unstable short trajectories that occur by chance but are not reliably reproducible under the current policy distribution, effectively trading task success for superficial step savings. This reveals that the step reduction objective alone is insufficient, as the policy can find shorter paths but collapses onto fragile shortcuts that fail to generalize.

\paragraph{Effect of Group-Anchored Regularization.} 
\begin{wrapfigure}{r}{0.46\textwidth}
  \centering
  \includegraphics[width=\linewidth]{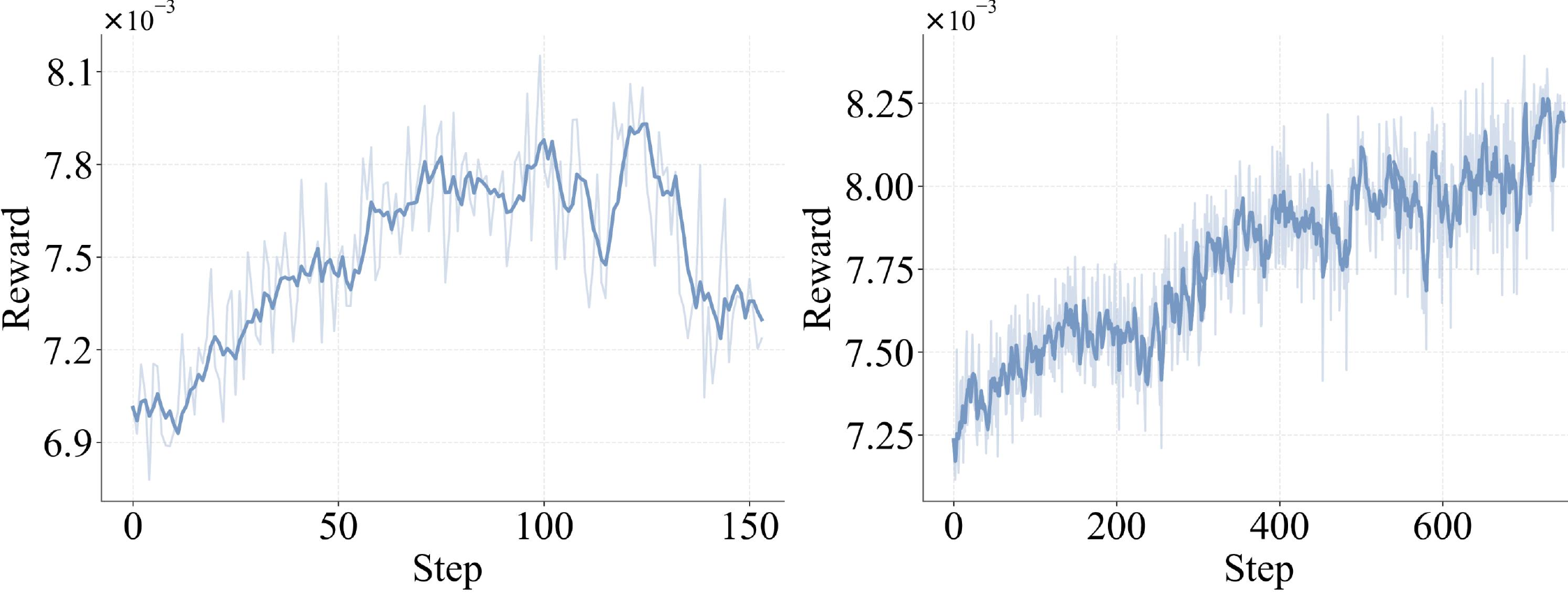}
  \captionsetup{width=\linewidth}
  \emergencystretch=5em
  \caption{Training reward curves without (Left) and with (Right) Group-Anchored Regularization on LIBERO-Spatial ($\pi_{0.5}$).}
  \label{fig:reward}
\end{wrapfigure}
 When Group-Anchored Regularization is added on top of the Step-Saving Reward, a striking improvement emerges. The success rate recovers from $93.7\%$ to $97.5\%$ while $S_{\text{total}}$ is further reduced from $81.7$ to $61.6$. As shown in Fig.~\ref{fig:reward}, without Group-Anchored Regularization, the reward collapses around step 125 as the policy exploits fragile short trajectories that cannot be reliably reproduced, causing training to destabilize. Incorporating Group-Anchored Regularization instead yields a steadily increasing reward throughout training. The fact that a regularization term simultaneously improves both task competence and execution efficiency reveals its essential role in steering the policy away from fragile shortcuts and toward genuinely concise and reproducible execution paths.

\paragraph{Effect of Dynamic Execution Horizon Exploration.} As reported in Table~\ref{tab:ablation_object}, fixing the acceptance ratio to a single value exposes a clear trade-off: as $\gamma$ increases, the execution window grows at the cost of progressive SR degradation, with Fixed $\gamma\!=\!0.75$ and $\gamma\!=\!1.0$ dropping to $95.8\%$ and $94.4\%$ respectively. Dynamic Execution Horizon Exploration resolves this tension by simultaneously probing multiple horizons within each group, achieving $h_{\text{chunk}}\!=\!15$ comparable to Fixed $\gamma\!=\!0.75$ while preserving an SR of $98.8\%$ nearly identical to the baseline, underscoring its critical role in stable and effective chunk extension.

\begin{table}[!t]
\centering
\caption{Combining PolicyTrim with VLA-Cache on the four LIBERO subsets using OpenVLA-OFT. PolicyTrim and VLA-Cache target orthogonal efficiency axes and yield compounded speedups.}
\label{tab:libero_openvla}
\setlength{\tabcolsep}{4pt}
\begin{tabular}{l cccc cccc}
\toprule
\multirow{2}{*}{Method} & \multicolumn{4}{c}{Spatial} & \multicolumn{4}{c}{Object} \\
\cmidrule(lr){2-5} \cmidrule(lr){6-9}
& SR & $S_{\text{total}}$ & $h_{\text{chunk}}$ & Spd$\uparrow$ & SR & $S_{\text{total}}$ & $h_{\text{chunk}}$ & Spd$\uparrow$ \\
\midrule
Baseline          & 97.8 & 113.1 & 8 & 1.0 & 97.6 & 138.8 & 8 & 1.0 \\
VLA-Cache         & 98.3 & -- & 8 & 1.26 & 97.5 & -- & 8 & 1.26 \\
$\mathbf{+}$PolicyTrim & 98.8 & 63.2 & 8 & 2.26 & 98.5 & 70.5 & 8 & 2.48 \\
\midrule
\midrule
\multirow{2}{*}{Method} & \multicolumn{4}{c}{Goal} & \multicolumn{4}{c}{Long} \\
\cmidrule(lr){2-5} \cmidrule(lr){6-9}
& SR & $S_{\text{total}}$ & $h_{\text{chunk}}$ & Spd$\uparrow$ & SR & $S_{\text{total}}$ & $h_{\text{chunk}}$ & Spd$\uparrow$ \\
\midrule
Baseline          & 97.6 & 115.7 & 8 & 1.0 & 94.2 & 256.3 & 8 & 1.0 \\
VLA-Cache         & 98.3 & - & 8 & 1.26 & 95.4 & - & 8 & 1.26 \\
$\mathbf{+}$PolicyTrim & 98.5 & 65.4 & 8 & 2.23 & 95.4 & 183.1 & 8 & 1.76 \\
\bottomrule
\end{tabular}
\end{table}

\subsection{Discussion}
\paragraph{What Causes Policy Inefficiency in VLA Policies?} The root cause likely lies in the training paradigms themselves. Imitation learning~\cite{liu2024rdt,octo_rss2024,spatialvla_rss2025} optimizes the policy to reproduce demonstrated behaviors without any explicit signal favoring execution efficiency. Moreover, prediction errors accumulate along action chunks due to distribution shift, causing the policy to take redundant corrective actions that further inflate the total execution steps.
 Standard RL post-training~\cite{tan2025piptvla,chen2025pi_} with binary success rewards also provides no explicit incentive for execution efficiency. Moreover, as the policy matures and success rates rise, reward variance within each sampled group tends to collapse, gradually diminishing the discriminative power of advantage estimates. 

\paragraph{Orthogonality with Compute-centric Methods.}
Policy efficiency and computational efficiency are orthogonal and jointly exploitable. While compute-centric methods reduce per-step inference latency, PolicyTrim targets the total number of forward inference calls, a dimension existing acceleration techniques leave entirely unaddressed. As demonstrated in Table~\ref{tab:libero_openvla}, integrating PolicyTrim with VLA-Cache yields speedups of up to 2.48$\times$ on LIBERO-Object, well beyond the 1.26$\times$ achievable by VLA-Cache alone, confirming that the two approaches together constitute a more complete path toward practical deployment.

%% file: sec/conclusion.tex
\section{Conclusion}
In this work, we identified policy efficiency as a critical yet overlooked bottleneck for deploying VLA models, and proposed \textbf{PolicyTrim}, a two-stage RL post-training framework that addresses this without architectural modifications or additional demonstrations. The first stage employs a dynamic horizon exploration mechanism to progressively push the trustworthy prediction frontier toward its empirical limit, while the second introduces a redundancy-aware reward coupled with group-anchored stability regularization to drive the policy toward genuinely concise execution without collapsing onto irreproducible shortcuts. Extensive experiments across diverse benchmarks and backbones demonstrate that PolicyTrim significantly reduces inference frequency while maintaining strong task competence. As an orthogonal complement to compute-centric acceleration, PolicyTrim can be seamlessly combined with existing inference optimization techniques for compounded efficiency gains, pointing toward a more holistic path to practical and scalable robotic deployments. While PolicyTrim provides a robust execution framework, future work may explore integrating PolicyTrim with on-device continuous learning to further adapt execution efficiency to dynamic and unseen environments.

\section*{Acknowledgment}
This work is supported by the National Natural Science Foundation of China (U23B2013, U2441242 and 62276176). This work was also partly supported by the SICHUAN Provincial Natural Science Foundation (No. 2024NSFJQ0023).

%% file: sec/sup.tex
\section{PolicyTrim Training Algorithm}
\subsection{PolicyTrim Training Algorithm}
We summarize the two-stage optimization procedure of PolicyTrim below. PolicyTrim improves intrinsic policy efficiency through two sequential GRPO-based post-training stages.

\input{algo}

Across all stages, each reward term is rescaled to the range of $r_{\mathrm{succ}}$. In Stage~2, the combined reward is renormalized after aggregation to preserve this range. Rewards are then standardized within each sampled group to compute group-relative advantages, and the policy is optimized with a clipped KL-regularized GRPO objective.

\section{Implementation Details}

All experiments used critic-free GRPO. We applied group-relative reward normalization and updated the policy directly from rollout returns, without a critic or separate reward model. Unless otherwise noted, training was conducted on a single node, using a GRPO group size of 8 throughout.

PolicyTrim has two sequential stages. Stage~1 extends reliable action chunks. Stage~2 reduces redundant steps using a step-saving reward and group-anchored stability regularization. Unless otherwise specified, their default coefficients were 0.8 and 0.2, respectively. Pi0.5 and GR00T used both stages, whereas OpenVLA-OFT used only Stage~2 due to backbone constraints. Each stage was trained for up to 500 epochs.

\subsection{Training Hyperparameters}

\noindent\textbf{Pi0.5.}
The global batch size was 2048 on LIBERO. The environment horizon was 160 for Spatial and Object, 240 for Goal, and 400 for LIBERO-10. The prediction horizon was fixed to 15 on LIBERO. The action horizon was set to [5, 10, 15] for Spatial, Goal, and Object, and to [10, 15] for LIBERO-10. On ManiSkill, the global batch size was 5120, the environment horizon was 80, the prediction horizon was 10, and the action horizon was [5, 10]. On MetaWorld, the global batch size was 2048, the environment horizon was 100, the prediction horizon was 10, and the action horizon was [5, 10].

\noindent\textbf{OpenVLA-OFT.}
The global batch size was 16384 on LIBERO. The environment horizon was 256 for Spatial, Object, and Goal, and 512 for LIBERO-10. The prediction horizon was fixed to 8, with action horizon [8]. On ManiSkill, the global batch size was 640, the environment horizon was 80, and both the prediction horizon and action horizon were fixed to 8.

\noindent\textbf{GR00T.}
The global batch size was 1024 throughout. On LIBERO-Spatial, the environment horizon was 160, with prediction horizon 10 and action horizon [5, 10]. On LIBERO-Object, Goal, and LIBERO-10, the prediction horizon was fixed to 10 and the action horizon to [10], with environment horizons of 160, 240, and 320, respectively.

\subsection{Backbone-Specific Settings}

\noindent\textbf{Pi0.5.}
Pi0.5 used the OpenPI backbone with FSDP in no-shard mode and without gradient checkpointing. All runs used three denoising steps. The action head used Flow-SDE, and the learning rate was $5\times10^{-6}$.

\noindent\textbf{OpenVLA-OFT.}
OpenVLA-OFT used the implementation in bfloat16 precision with full-parameter fine-tuning. All checked-in GRPO configurations enabled gradient checkpointing. The learning rate was $2\times10^{-5}$.

\noindent\textbf{GR00T.}
GR00T used bfloat16 precision and four denoising steps. Gradient checkpointing was disabled. The learning rate was $5\times10^{-6}$.

\begin{figure}[!t]
  \centering
  \includegraphics[width=\textwidth]{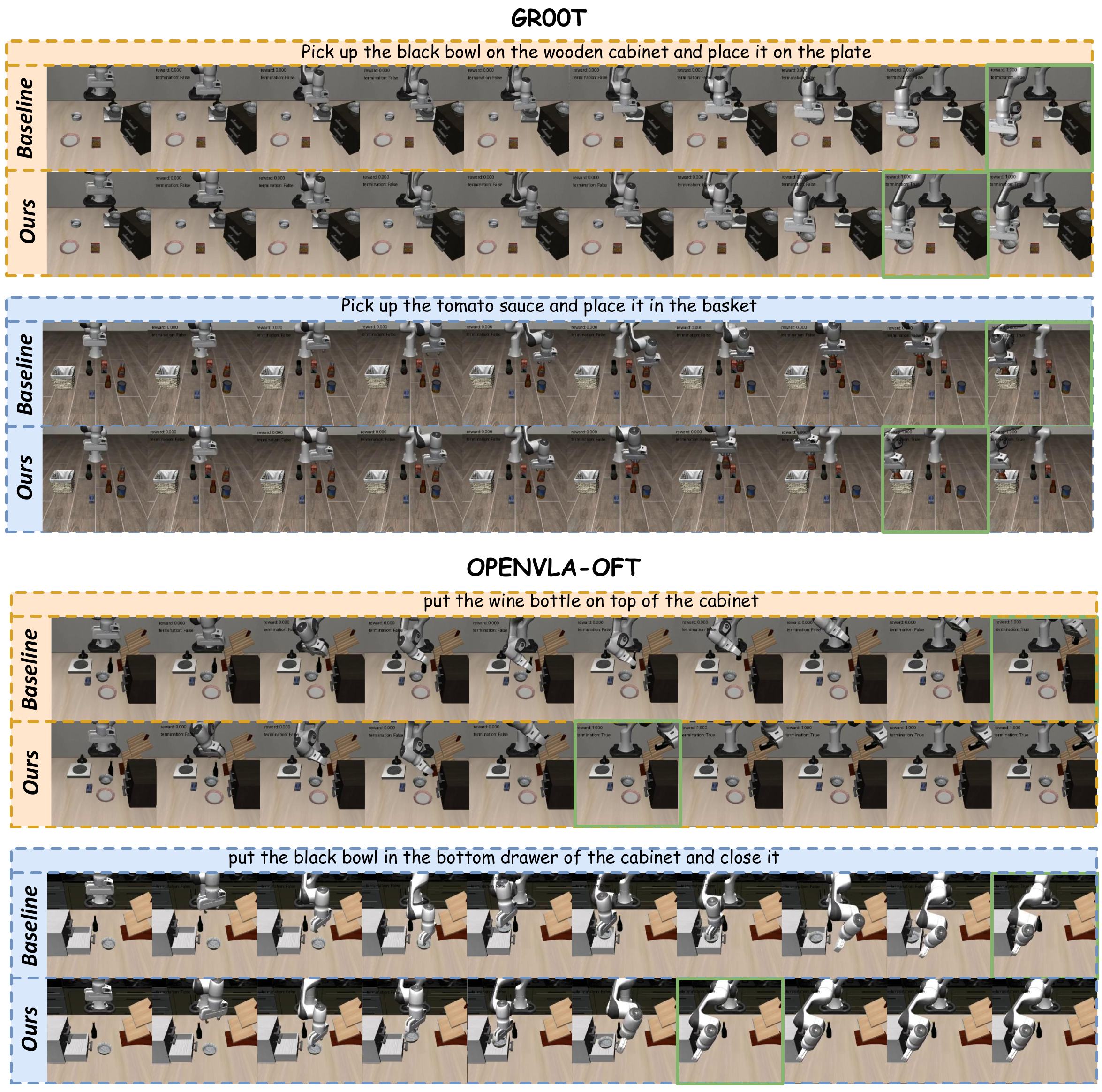}
  \caption{\textbf{Qualitative comparisons on GR00T and OpenVLA-OFT.}
For each instruction, we compare execution snapshots of the baseline policy and PolicyTrim. Across both backbones, PolicyTrim reaches the goal with more compact trajectories and fewer redundant physical steps. Green boxes highlight salient late-stage frames where task completion occurs.}
\label{fig:qualitative_gr00t_openvla}
\end{figure}
\section{Additional Results}

\subsection{Qualitative Results}

As shown in Fig.~\ref{fig:qualitative_gr00t_openvla}, PolicyTrim consistently reaches the target state with more compact execution trajectories across both backbones. On GR00T, the optimized policy preserves reliable long-horizon behavior while reducing unnecessary motion before successful placement. On OpenVLA-OFT, PolicyTrim produces noticeably shorter and cleaner trajectories, especially in tasks that otherwise involve hesitation or redundant adjustments. These examples qualitatively support our main claim that PolicyTrim improves intrinsic policy efficiency by simultaneously enlarging the reliable action horizon and reducing redundant physical steps.

\subsection{Ablation on Group Size}

\begin{table*}[h]
\centering
\caption{Ablation on group size $G$ for $\pi_{0.5}$ on the four LIBERO subsets. We report success rate (SR), average physical steps ($S_{\text{total}}$), average action chunk execution length ($h_{\text{chunk}}$), and end-to-end execution speedup (Spd).}
\label{tab:group_ablation_pi05}
\resizebox{\textwidth}{!}{
\begin{tabular}{c cccc cccc cccc cccc}
\toprule
\multirow{2}{*}{$G$}
& \multicolumn{4}{c}{Spatial}
& \multicolumn{4}{c}{Object}
& \multicolumn{4}{c}{Goal}
& \multicolumn{4}{c}{Long} \\
\cmidrule(lr){2-5} \cmidrule(lr){6-9} \cmidrule(lr){10-13} \cmidrule(lr){14-17}
& SR & $S_{\text{total}}$ & $h_{\text{chunk}}$ & Spd$\uparrow$
& SR & $S_{\text{total}}$ & $h_{\text{chunk}}$ & Spd$\uparrow$
& SR & $S_{\text{total}}$ & $h_{\text{chunk}}$ & Spd$\uparrow$
& SR & $S_{\text{total}}$ & $h_{\text{chunk}}$ & Spd$\uparrow$ \\
\midrule
8  & 97.8 & 59.8 & 15 & 5.43 & 98.5 & 64.3 & 15 & 5.83 & 98.8 & 63.5 & 15 & 5.23 & 93.3 & 171.8 & 10 & 2.91 \\
10 & 97.4 & 60.5 & 15 & 5.37 & 98.1 & 65.0 & 15 & 5.77 & 98.4 & 64.2 & 15 & 5.17 & 92.9 & 172.6 & 10 & 2.89 \\
12 & 97.9 & 59.4 & 15 & 5.47 & 98.6 & 63.9 & 15 & 5.87 & 98.9 & 63.1 & 15 & 5.26 & 93.5 & 171.2 & 10 & 2.92 \\
14 & 97.6 & 60.1 & 15 & 5.41 & 98.3 & 64.7 & 15 & 5.80 & 98.6 & 63.8 & 15 & 5.20 & 93.1 & 172.1 & 10 & 2.90 \\
16 & 97.3 & 60.7 & 15 & 5.35 & 98.0 & 65.1 & 15 & 5.76 & 98.2 & 64.4 & 15 & 5.15 & 92.7 & 172.7 & 10 & 2.89 \\
\bottomrule
\end{tabular}
}
\end{table*}

\begin{table*}[h]
\centering
\caption{Ablation study of different PolicyTrim components on the four LIBERO subsets using $\pi_{0.5}$.}
\label{tab:ablation_components_all}
\renewcommand{\arraystretch}{0.95}
\resizebox{\textwidth}{!}{
\begin{tabular}{ccc cccc cccc}
\toprule
\multirow{2}{*}{\makecell[c]{Reliable Chunk\\Extension}} &
\multirow{2}{*}{\makecell[c]{Step-Saving\\Reward}} &
\multirow{2}{*}{\makecell[c]{Group-Anchored\\Regularization}} &
\multicolumn{4}{c}{Spatial} &
\multicolumn{4}{c}{Object} \\
\cmidrule(lr){4-7} \cmidrule(lr){8-11}
& & &
SR & $S_{\text{total}}$ & $h_{\text{chunk}}$ & Spd$\uparrow$ &
SR & $S_{\text{total}}$ & $h_{\text{chunk}}$ & Spd$\uparrow$ \\
\midrule
\multicolumn{1}{c}{\textcolor[gray]{0.8}{\ding{55}}} &
\multicolumn{1}{c}{\textcolor[gray]{0.8}{\ding{55}}} &
\multicolumn{1}{c}{\textcolor[gray]{0.8}{\ding{55}}}
& 97.8 & 108.3 & 5  & 1.00
& 99.1 & 125.0 & 5  & 1.00 \\
\multicolumn{1}{c}{\ding{51}} &
\multicolumn{1}{c}{\textcolor[gray]{0.8}{\ding{55}}} &
\multicolumn{1}{c}{\textcolor[gray]{0.8}{\ding{55}}}
& 97.4 & 111.6 & 15 & 2.91
& 98.8 & 128.6 & 15 & 2.92 \\
\multicolumn{1}{c}{\textcolor[gray]{0.8}{\ding{55}}} &
\multicolumn{1}{c}{\ding{51}} &
\multicolumn{1}{c}{\textcolor[gray]{0.8}{\ding{55}}}
& 95.1 & 86.9 & 5  & 1.25
& 96.7 & 99.8 & 5  & 1.25 \\
\multicolumn{1}{c}{\textcolor[gray]{0.8}{\ding{55}}} &
\multicolumn{1}{c}{\ding{51}} &
\multicolumn{1}{c}{\ding{51}}
& 97.1 & 65.5 & 5  & 1.65
& 98.4 & 72.6 & 5  & 1.72 \\
\multicolumn{1}{c}{\ding{51}} &
\multicolumn{1}{c}{\ding{51}} &
\multicolumn{1}{c}{\ding{51}}
& 97.8 & 59.8 & 15 & 5.43
& 98.5 & 64.3 & 15 & 5.83 \\
\midrule
\multirow{2}{*}{\makecell[c]{Reliable Chunk\\Extension}} &
\multirow{2}{*}{\makecell[c]{Step-Saving\\Reward}} &
\multirow{2}{*}{\makecell[c]{Group-Anchored\\Regularization}} &
\multicolumn{4}{c}{Goal} &
\multicolumn{4}{c}{Long} \\
\cmidrule(lr){4-7} \cmidrule(lr){8-11}
& & &
SR & $S_{\text{total}}$ & $h_{\text{chunk}}$ & Spd$\uparrow$ &
SR & $S_{\text{total}}$ & $h_{\text{chunk}}$ & Spd$\uparrow$ \\
\midrule
\multicolumn{1}{c}{\textcolor[gray]{0.8}{\ding{55}}} &
\multicolumn{1}{c}{\textcolor[gray]{0.8}{\ding{55}}} &
\multicolumn{1}{c}{\textcolor[gray]{0.8}{\ding{55}}}
& 98.7 & 110.6 & 5  & 1.00
& 93.0 & 249.8 & 5  & 1.00 \\
\multicolumn{1}{c}{\ding{51}} &
\multicolumn{1}{c}{\textcolor[gray]{0.8}{\ding{55}}} &
\multicolumn{1}{c}{\textcolor[gray]{0.8}{\ding{55}}}
& 98.3 & 113.9 & 15 & 2.91
& 92.6 & 253.4 & 10 & 1.97 \\
\multicolumn{1}{c}{\textcolor[gray]{0.8}{\ding{55}}} &
\multicolumn{1}{c}{\ding{51}} &
\multicolumn{1}{c}{\textcolor[gray]{0.8}{\ding{55}}}
& 95.8 & 88.9 & 5  & 1.24
& 89.8 & 205.2 & 5  & 1.22 \\
\multicolumn{1}{c}{\textcolor[gray]{0.8}{\ding{55}}} &
\multicolumn{1}{c}{\ding{51}} &
\multicolumn{1}{c}{\ding{51}}
& 97.7 & 68.9 & 5  & 1.61
& 91.9 & 191.3 & 5  & 1.31 \\
\multicolumn{1}{c}{\ding{51}} &
\multicolumn{1}{c}{\ding{51}} &
\multicolumn{1}{c}{\ding{51}}
& 98.8 & 63.5 & 15 & 5.23
& 93.3 & 171.8 & 10 & 2.91 \\
\bottomrule
\end{tabular}
}
\end{table*}

The results show that PolicyTrim is largely insensitive to the choice of group size $G$. As $G$ varies from 8 to 16, all four LIBERO subsets exhibit only minor fluctuations in success rate, total execution steps, and end-to-end speedup, without any consistent trend. These results suggest that the default choice $G=8$ is already sufficient in practice.

\subsection{Ablation on Components}
In Table~\ref{tab:ablation_components_all}, we extend the component ablation study to all four LIBERO subsets using the $\pi_{0.5}$ model. The results show a consistent pattern across all subsets. Reliable Action Chunk Extension alone mainly improves execution speed by enlarging the action horizon, but slightly increases the number of physical steps and can also lead to a small drop in success rate. In contrast, the Step-Saving Reward alone reduces $S_{\text{total}}$ substantially, but this comes at the cost of degraded task success, indicating that shorter trajectories alone are not necessarily reliable. Adding Group-Anchored Regularization largely recovers task competence while further improving execution efficiency, showing its role in stabilizing concise behaviors. Finally, enabling all three components together gives the best overall trade-off, consistently achieving the highest end-to-end speedup while preserving strong task performance across Spatial, Object, Goal, and Long.

\subsection{Real-World Deployment}
\label{subsec:real_world_deployment}

\begin{figure}[h]
  \centering
  \includegraphics[width=\textwidth]{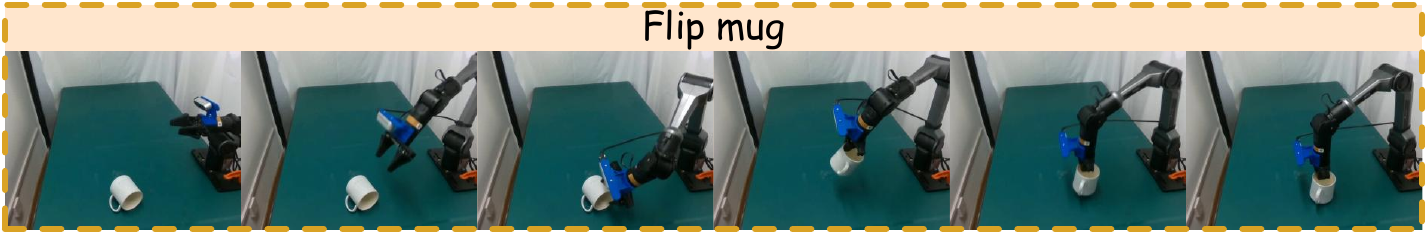}
  \caption{Real-world execution visualization on the FlipMug task.}
  \label{fig:app_real_world}
\end{figure}

We deploy PolicyTrim on a real robot platform and evaluate it on three
tabletop manipulation tasks: FlipMug, HangMug, and TapeBox.
Figure~\ref{fig:app_real_world} shows the real-world execution
visualization on the FlipMug task. More real-world visualizations are
available on the project GitHub page.

\subsection{Robustness under Visual Perturbations}
\label{subsec:ood_robustness}

We further evaluate PolicyTrim under visual distribution shifts in
simulation. Specifically, we test on LIBERO-Spatial with two visual
perturbations: Gaussian blur with kernel size $k=13$ and 50\% random
occlusion. These perturbations evaluate whether the learned efficient
execution strategy remains robust when visual observations become
degraded or partially missing.

\begin{table}[h]
\centering
\caption{Simulation robustness results on LIBERO-Spatial under visual perturbations. We report SR / Step, where SR is success rate in \% and Step is the average number of physical execution steps.}
\label{tab:app_ood}
\begin{tabular}{lcc}
\toprule
Method & Gaussian Blur & Random Occlusion \\
\midrule
Baseline   & 79.6 / 128.9 & 83.2 / 120.3 \\
PolicyTrim & 88.4 /  77.6 & 90.6 /  75.9 \\
\bottomrule
\end{tabular}
\end{table}

As shown in Table~\ref{tab:app_ood}, PolicyTrim consistently outperforms
the baseline under both perturbations. Besides improving success rate, it
also substantially reduces the number of physical steps, indicating that
PolicyTrim does not merely overfit to clean visual observations but
learns a more robust and concise execution strategy.

\subsection{Post-Training Cost}
\label{subsec:post_training_cost}

PolicyTrim is built on top of the RLinf framework and follows its rollout
and environment-interaction settings for each task. The total
post-training cost is approximately 68 hours wall-clock time on
8$\times$H100 GPUs, compared with approximately 78 hours for the
corresponding RLinf setting. The reduced training time is mainly because
the extended reliable action chunk horizon reduces the number of
inference calls required per rollout. Since the learned policy-efficiency
improvement generalizes across tasks and can be reused after
post-training, this one-time training cost is justified by the resulting
deployment-time speedup.

\subsection{Horizon-Sweep Baseline}
\label{subsec:horizon_sweep}

To verify that PolicyTrim's improvement does not simply come from
executing longer action chunks, we compare against fixed-horizon
execution baselines. For $\pi_{0.5}$, we evaluate different fixed
execution horizons $h$ and report the resulting success rate and physical
step count.

\begin{table}[h]
\centering
\caption{Horizon-sweep baseline for $\pi_{0.5}$. Fixed larger horizons degrade success rate, while PolicyTrim learns to extend the reliable horizon through RL post-training.}
\label{tab:app_horizon_sweep}
\begin{tabular}{lcc}
\toprule
Method & SR & Step \\
\midrule
Fixed $h=5$  & 97.8 & 108.3 \\
Fixed $h=10$ & 97.2 & 109.5 \\
Fixed $h=15$ & 94.1 & 110.2 \\
Fixed $h=20$ & 93.1 & 117.5 \\
\midrule
Stage~1 only & 97.2 & 111.8 \\
Full PolicyTrim & \textbf{97.8} & \textbf{59.8} \\
\bottomrule
\end{tabular}
\end{table}

As shown in Table~\ref{tab:app_horizon_sweep}, naively increasing the
fixed execution horizon causes the success rate to drop from 97.8\% to
93.1\%. In contrast, Stage~1 maintains a high success rate under an
extended action horizon, confirming that the gain comes from RL-based
reliable chunk extension rather than from simply forcing the model to
execute longer chunks. The full PolicyTrim pipeline further reduces the
physical step count from 111.8 to 59.8 through Stage~2 step reduction.

\subsection{Hyperparameter Sensitivity}
\label{subsec:hyperparameter_sensitivity}

We evaluate the sensitivity of PolicyTrim to three key hyperparameters:
the step-budget multiplier $\alpha$, the GRPO group size $G$, and the
stability regularization coefficient $\lambda_{\text{stab}}$. All
experiments are conducted on LIBERO-Spatial using $\pi_{0.5}$.

\begin{table}[h]
\centering
\caption{Hyperparameter sensitivity on LIBERO-Spatial. We report SR / Step. Default values are shown in bold.}
\label{tab:app_hyperparameter}
\begin{tabular}{rlccccc}
\toprule
\multirow{2}{*}{$\alpha$}
 & Value   & 1.1 & 1.2 & \textbf{1.3} & 1.5 & 2.0 \\
 & SR/Step & 97.2/61.2 & 97.5/60.4 & \textbf{97.8/59.8} & 97.9/60.5 & 97.6/61.8 \\
\midrule
\multirow{2}{*}{$\lambda_{\text{stab}}$}
 & Value   & 0.05 & 0.10 & \textbf{0.15} & 0.20 & 0.30 \\
 & SR/Step & 97.1/58.9 & 97.5/59.3 & \textbf{97.8/59.8} & 97.7/60.4 & 97.2/61.2 \\
\bottomrule
\end{tabular}
\end{table}

The results in Table~\ref{tab:app_hyperparameter} show that PolicyTrim is
robust to these hyperparameters. Across all tested settings, the success
rate varies within a narrow range and the physical step count remains
stable. This indicates that the default configuration works well without
requiring careful per-task tuning.

\section{Failure Case Analysis}
We provide several representative failure cases for qualitative analysis. These examples highlight scenarios where the policy does not fully complete the task or requires additional corrections during execution.

\begin{figure}[h]
  \centering
  \includegraphics[width=\textwidth]{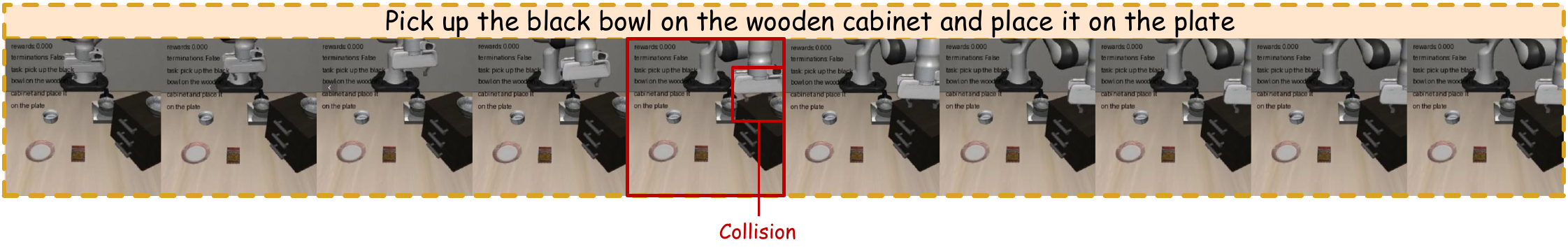}
  \caption{\textbf{Failure case without group-anchored stability regularization.} The policy approaches the bowl with insufficient clearance, causing a collision and task failure.}
\label{fig:Failure_Case}
\end{figure}

In this failure case, removing the group-anchored stability regularization causes the policy to overemphasize step-saving behavior and pursue execution speed too aggressively. To grasp the bowl on the wooden cabinet more quickly, the robot follows a shorter but unsafe trajectory, without lifting the end-effector high enough before approaching the target. As a result, the gripper collides with the bowl, knocking it out of the reachable workspace and leading to task failure. This example highlights the importance of group-anchored stability regularization in preventing overly aggressive efficiency optimization and preserving safer interaction behavior.

%% file: algo.tex
\begin{algorithm}[h]
\caption{PolicyTrim Post-Training}
\label{alg:policytrim}
\begin{algorithmic}[1]
\Require Policy $\pi_\theta$, reference policy $\pi_{\mathrm{ref}}\!\leftarrow\!\pi_\theta$, task distribution $\mathcal{T}$
\Require Acceptance ratios $\Gamma$, max chunk length $H$, group size $G$, step budget $S_{\mathrm{base}}$
\Require Success reward $r_{\mathrm{succ}}$, coefficients $\beta_{\mathrm{hor}}, \lambda_{\mathrm{stab}}, \beta_{\mathrm{KL}}$, constants $\sigma_{\mathrm{floor}}, \epsilon_{\mathrm{adv}}, \epsilon_{\mathrm{clip}}$

\Statex \textit{GRPO-Update}$(\pi_\theta,\{R_i\}_{i=1}^G)$:
\Statex \hspace{\algorithmicindent} $A_i \leftarrow (R_i-\mu_R)/(\sigma_R+\epsilon_{\mathrm{adv}})$, with $\mu_R,\sigma_R$ computed over $\{R_i\}_{i=1}^G$
\Statex \hspace{\algorithmicindent} Update $\pi_\theta$ with clipped KL-regularized GRPO using $\{A_i\}_{i=1}^G$

\Statex \textbf{Stage 1: Reliable Action Chunk Extension}
\For{each iteration}
    \State Sample $(l,o_0)\sim\mathcal{T}$
    \For{$i=1$ to $G$}
        \State Sample $\gamma_i\sim\Gamma$, set $h_i\leftarrow \lfloor \gamma_i H \rfloor$, and roll out $\tau_i$
        \State $R_i \leftarrow \mathcal{I}_{\mathrm{succ}}^{(i)} \bigl(r_{\mathrm{succ}}+\beta_{\mathrm{hor}}\gamma_i\bigr)$
    \EndFor
    \State \textit{GRPO-Update}$(\pi_\theta,\{R_i\}_{i=1}^G)$
\EndFor

\State $\pi_{\mathrm{ref}} \leftarrow \pi_\theta$

\Statex \textbf{Stage 2: Redundancy-Aware Step Reduction}
\For{each iteration}
    \State Sample $(l,o_0)\sim\mathcal{T}$ and set $h\leftarrow \lfloor \max(\Gamma)H \rfloor$
    \For{$i=1$ to $G$}
        \State Roll out $\tau_i$ with window $h$, and record $\mathcal{I}_{\mathrm{succ}}^{(i)}$ and $S_i$
    \EndFor
    \State Compute $\mu_{\mathrm{group}},\sigma_{\mathrm{group}}$ over successful $\{S_i\}$
    \For{$i=1$ to $G$}
        \State $R_i \leftarrow \mathcal{I}_{\mathrm{succ}}^{(i)}
        \Bigl(r_{\mathrm{succ}}+\frac{\max(0,S_{\mathrm{base}}-S_i)}{S_{\mathrm{base}}}
        -\lambda_{\mathrm{stab}}\tanh\!\bigl(\frac{|S_i-\mu_{\mathrm{group}}|}{\max(\sigma_{\mathrm{group}},\sigma_{\mathrm{floor}})}\bigr)\Bigr)$
    \EndFor
    \State \textit{GRPO-Update}$(\pi_\theta,\{R_i\}_{i=1}^G)$
\EndFor

\State \Return $\pi_\theta$
\end{algorithmic}
\end{algorithm}